\documentclass[letterpaper]{article} 
\usepackage{aaai24}  
\usepackage{times}  
\usepackage{helvet}  
\usepackage{courier}  
\usepackage[hyphens]{url}  
\usepackage{graphicx} 
\usepackage{natbib}  
\usepackage{caption} 
\usepackage{algorithm}
\usepackage{algorithmic}
\usepackage{subcaption}

\usepackage{newfloat}
\usepackage{listings}
\DeclareCaptionStyle{ruled}{labelfont=normalfont,labelsep=colon,strut=off} 
\floatstyle{ruled}
\newfloat{listing}{tb}{lst}{}
\floatname{listing}{Listing}
\title{Watch Your Head: Assembling Projection Heads to Save the Reliability of Federated Models}
\author {
    Jinqian Chen\textsuperscript{\rm 1, 3},
    Jihua Zhu\textsuperscript{\rm 1}\thanks{Corresponding Author}\textsuperscript{\rm, 3},
    Qinghai Zheng\textsuperscript{\rm 2},
    Zhongyu Li\textsuperscript{\rm 1, 3},
    Zhiqiang Tian\textsuperscript{\rm 1, 3}
}
\affiliations {
    \textsuperscript{\rm 1}School of Software Engineering, Xi'an Jiaotong University\\
    \textsuperscript{\rm 2}College of Computer and Data Science, Fuzhou University\\
    \textsuperscript{\rm 3}Shaanxi Joint Key Laboratory for Artificial Intelligence, China\\
    chenjinqian@stu.xjtu.edu.cn, zhujh@xjtu.edu.cn,
    zhengqinghai@fzu.edu.cn
}

\usepackage{algorithm}
\usepackage{algorithmic}
\usepackage{amsmath}
\usepackage{amssymb}
\usepackage{amsthm}

\newtheorem{definition}{Definition}
\usepackage{subcaption}
\usepackage{booktabs}
\usepackage{multirow}

\begin{document}

\maketitle

\begin{abstract}
Federated learning encounters substantial challenges with heterogeneous data, leading to performance degradation and convergence issues. While considerable progress has been achieved in mitigating such an impact, the reliability aspect of federated models has been largely disregarded. In this study, we conduct extensive experiments to investigate the reliability of both generic and personalized federated models. Our exploration uncovers a significant finding: \textbf{federated models exhibit unreliability when faced with heterogeneous data}, demonstrating poor calibration on in-distribution test data and low uncertainty levels on out-of-distribution data. This unreliability is primarily attributed to the presence of biased projection heads, which introduce miscalibration into the federated models. Inspired by this observation, we propose the "Assembled Projection Heads" (APH) method for enhancing the reliability of federated models. By treating the existing projection head parameters as priors, APH randomly samples multiple initialized parameters of projection heads from the prior and further performs targeted fine-tuning on locally available data under varying learning rates. Such a head ensemble introduces parameter diversity into the deterministic model, eliminating the bias and producing reliable predictions via head averaging. We evaluate the effectiveness of the proposed APH method across three prominent federated benchmarks. Experimental results validate the efficacy of APH in model calibration and uncertainty estimation. Notably, APH can be seamlessly integrated into various federated approaches but only requires less than 30\% additional computation cost for 100$\times$ inferences within large models.
\end{abstract}

\section{Introduction}

Federated learning is a training paradigm that holds promise for privacy preservation \cite{mcmahan2017communication, wei2020federated}. This approach does not require the central server to collect clients' data. Instead, clients train their local models and upload the parameters to the server for aggregation. In addition to privacy concerns, the reliability of neural networks has also garnered considerable attention, given their deployment in numerous critical scenarios \cite{levinson2011towards, miotto2016deep}. Recent research has shown that modern neural networks tend to exhibit overconfidence \cite{guo2017calibration}. This issue is far from trivial, especially in classification networks, where overconfidence can lead to higher predicted confidence in the sample than the actual probability of its assigned class. Even for the out-of-distribution sample, the model will assign it to a specific class with high confidence (i.e. low uncertainty) attached. Such discrepancies have grave implications for decision-making and significantly harms the model's reliability, rendering softmax outputs unsuitable as uncertainty indicators.

\begin{figure}[t]
    \centering
    \includegraphics[width=0.45\textwidth]{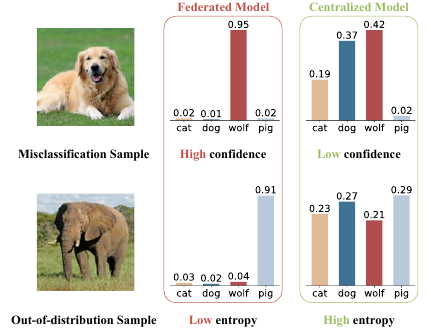}
    \caption{{Generic Federated Models are Not Reliable}. Compared with the centralized training models, the generic federated models tend to be more overconfident on misclassified samples and exhibit lower uncertainty (i.e. lower predictive entropy) on out-of-distribution samples (See Section 3), demonstrating the serious reliability issue.}
    \label{fig:enter-label}
\end{figure}
Privacy protection and reliability guarantees have a large cross field in practical application scenarios. Taking smart healthcare for example, it's always impractical to collect the private data of patients to a central server for the training of a diagnostic classification model, and thus leads to the broad application of federated learning. However, privacy is not all we are after. In such important application scenarios, it is natural to chase for the reliability of the model, expecting it to output low confidence in misclassified diagnoses and refuses to make decisions with unknown diagnoses. So here comes the natural question: \textbf{\textit{Whether the federated model is reliable? Is it well-calibrated and sensitive to the out-of-distribution data?}}

Unfortunately, the answer to this question still remains unclear. Despite the numerous challenges associated with federated learning \cite{kairouz2021advances, li2020federatedSurvey}, the issue of reliability has received less attention, although the problem could be even worse (refer to Section 3). In federated learning, the complex data distribution among different clients often makes the basic I.I.D assumption invalid, resulting in poor convergence and performance degradation of the global model. To address these issues, a great number of methods have been proposed to alleviate the impact of the Non-IID data \cite{ karimireddy2020scaffold, li2021model, li2020federated}. However, none of these methods attempts to assess how the federated framework affects the model's reliability, much less improve it.

What's worse, most existing calibration and uncertainty estimation methods, which can improve the reliability of models, can not be applied to the federated models directly. 
MCDropout \cite{gal2016dropout} requires the presence of dropout layers in the network, which are rarely used in current network designs because of the discrepancy between dropout and BatchNorm layers \cite{li2019understanding}. Deep Ensembles \cite{riquelme2018deep} is also an empirical but effective method of uncertainty estimation. However, it requires training the networks with different random initializations multiple times, which is costly and impractical in federated learning. It is also impractical to convert the model structure into the Bayesian model for the application of the Bayesian-based uncertainty estimation methods. Except for these classic methods, the latest SOTA uncertainty estimation methods always require additional data operation \cite{deltauq}, updating of global class centroids \cite{DUQ}, well-trained checkpoints \cite{maddox2019simple}, etc, making these effective methods not applicable to federated frameworks. 

In this paper, by conducting extensive experiments on the popular benchmark dataset, we provide a systematic investigation of the reliability of the federated model. We uncover the fact that \textbf{the federated model is unreliable compared to the model of centralized training}. Generic federated models tend to be more miscalibrated while personalized federated models exhibit lower uncertainty (i.e. less insensitive) to the OOD data. Experimental results illustrate that data heterogeneity cooperated with partial participation significantly harms the model's reliability while the impact of other factors is trivial. We further demonstrate that the biased projection head is one of the main causes of the reliability degradation. Motivated by this observation, we proposed a lightweight but effective uncertainty estimation method named APH for federated learning to improve the federated model's reliability. By randomly permuting the obtained federated parameter of the projection head as initialization, APH fine-tunes multiple projection heads with various learning rates to explore its parameter space and introduce parameter diversity, producing reliable predictions through head assembling and averaging. 

The contributions of this paper are delivered as follows:\\
1) We provide a systematic analysis of the reliability of the federated model. We uncover the fact that the federated model is unreliable compared with the centralized training model and further investigate its impact factors. \\2) We propose a lightweight but effective federated uncertainty estimation method named APH. APH can be seamlessly integrated into most SOTA methods to improve the performance and reliability of federated models. \\3) We validate the effectiveness of APH on prominent federated benchmarks with models of various sizes, showing its effectiveness and efficiency in improving reliability.

\begin{figure}[t!]
\captionsetup[subfigure]{justification=centering}
    \centering
      \begin{subfigure}{0.23\textwidth}
        \includegraphics[width=\textwidth]{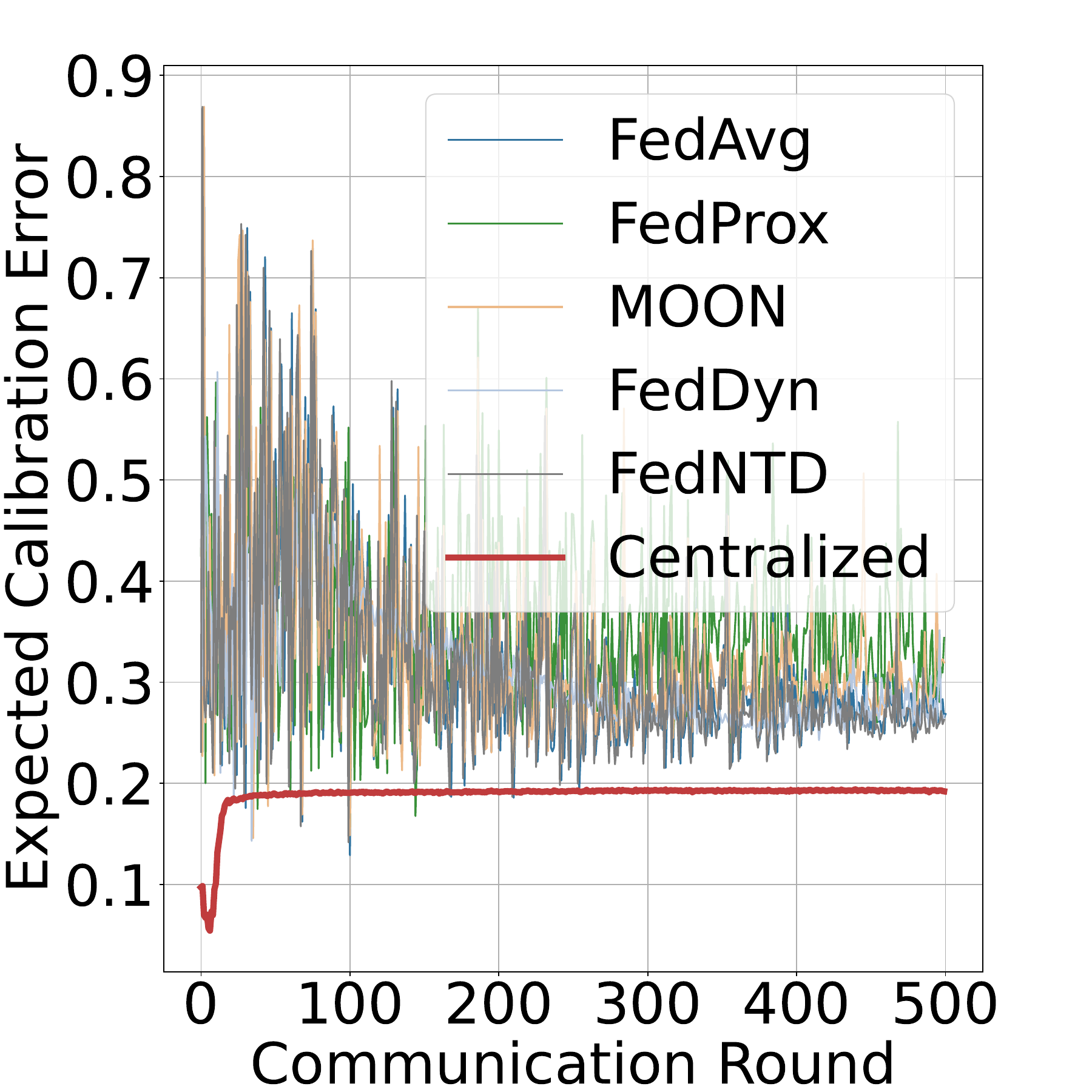}
          \caption{In-domain F-ECE}
          \label{fig2a}
      \end{subfigure}
      \begin{subfigure}{0.23\textwidth}
        \includegraphics[width=\textwidth]{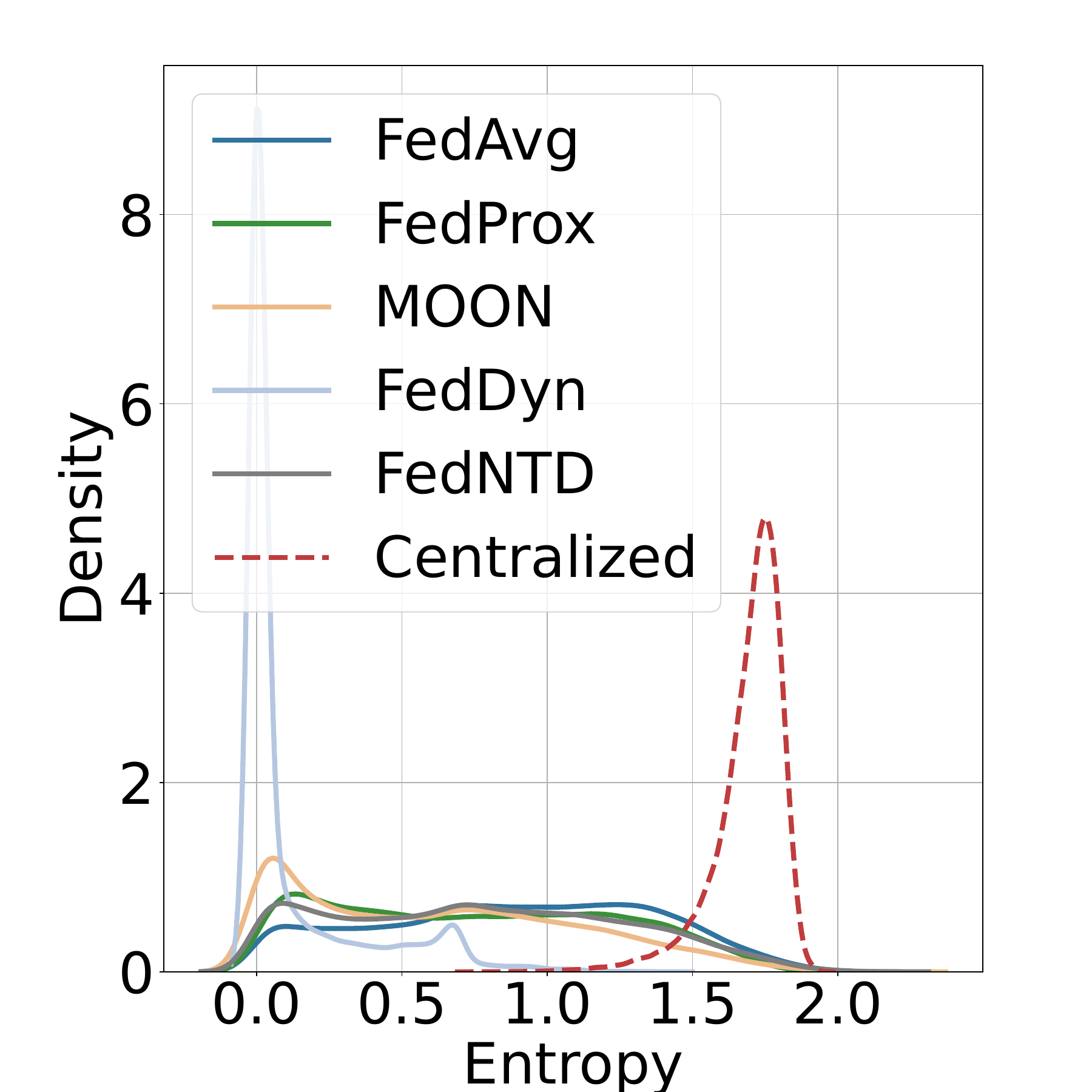}
          \caption{OOD Predictive Entropy}
          \label{fig2b}
      \end{subfigure}
    \caption{{Reliability of Generic Federated Models.} (a) F-ECE of different generic federated models compared with the centralized training model on in-domain test data. F-ECE of generic federated models is significantly higher than centralized training models, indicating severe overconfidence problems. (b) Histograms of predictive distribution entropy on OOD dataset. The predictive entropy of generic federated models is dramatically lower than the centralized training model, showing lower uncertainty levels to OOD samples.}
    \label{fig2}
\end{figure}

\section{Background and Related Work}

\subsection{Problem Setup}
We consider a practical horizontal federated scenario \cite{yang2019federated} with the Non-IID data distribution among clients. In this paper, we mainly focus on the skewness in label distribution and the quantity skewness \cite{li2022federated, zhu2021federated}. The skewness of the feature distribution is out of the scope of our paper as it is usually what vertical federated learning is concerned with. 

Assume that there are ${N}$ independent clients. Each client ${c_i}$ has their own local training data $\mathcal{D}_i = \left\{\left(\mathbf{x}_j^i, \mathbf{y}_j^i\right)\right\}_{j=1}^{n_i}$.  The aim of federated learning is to utilize this distributed dataset $\mathcal{D} = \{\mathcal{D}_1, \mathcal{D}_2, ... , \mathcal{D}_N\}$ to train a generic model $f\left(\cdot; \boldsymbol{\theta}_{g}\right)$ or a set of personalized models $\{f\left(\cdot; \boldsymbol{\theta}_{i}\right)\}_{i=1}^{N}$ in $\mathcal{R}$ communication rounds for the $K$-classification problem. We denote $\gamma$ as the participation ratio. In each communication round $r$, we first select $\lceil \gamma N \rceil$  clients and get the participated client set $\mathcal{B}_{r}$. For each client $c_i \in \mathcal{B}_{r}$, we distribute the current global model parameter $\boldsymbol{{\theta}}^{r-1}_{g}$ to client $c_i$, and update its local model parameter $\boldsymbol{{\theta}}^{r}_{i}$. Typically, in FedAvg, we set $\boldsymbol{{\theta}}^{r}_{i} = \boldsymbol{{\theta}}_{g}^{r-1}$. Then each client $c_i$ utilize its local data $\mathcal{D}_i$ to update its local model parameter $\boldsymbol{{\theta}}^{r}_{i}$ for $E$ epochs. All the updated model parameter $\hat{\boldsymbol{{\theta}}}^{r}_{i}$ of client $c_i$ in $\mathcal{B}_{r}$ will be uploaded to the server and used to get the global model $\boldsymbol{{\theta}}^{r}_{g}$.

\begin{figure}[t!]
    \captionsetup[subfigure]{justification=centering}
    \centering
    \begin{subfigure}{0.2\textwidth}
        \includegraphics[width=\textwidth]{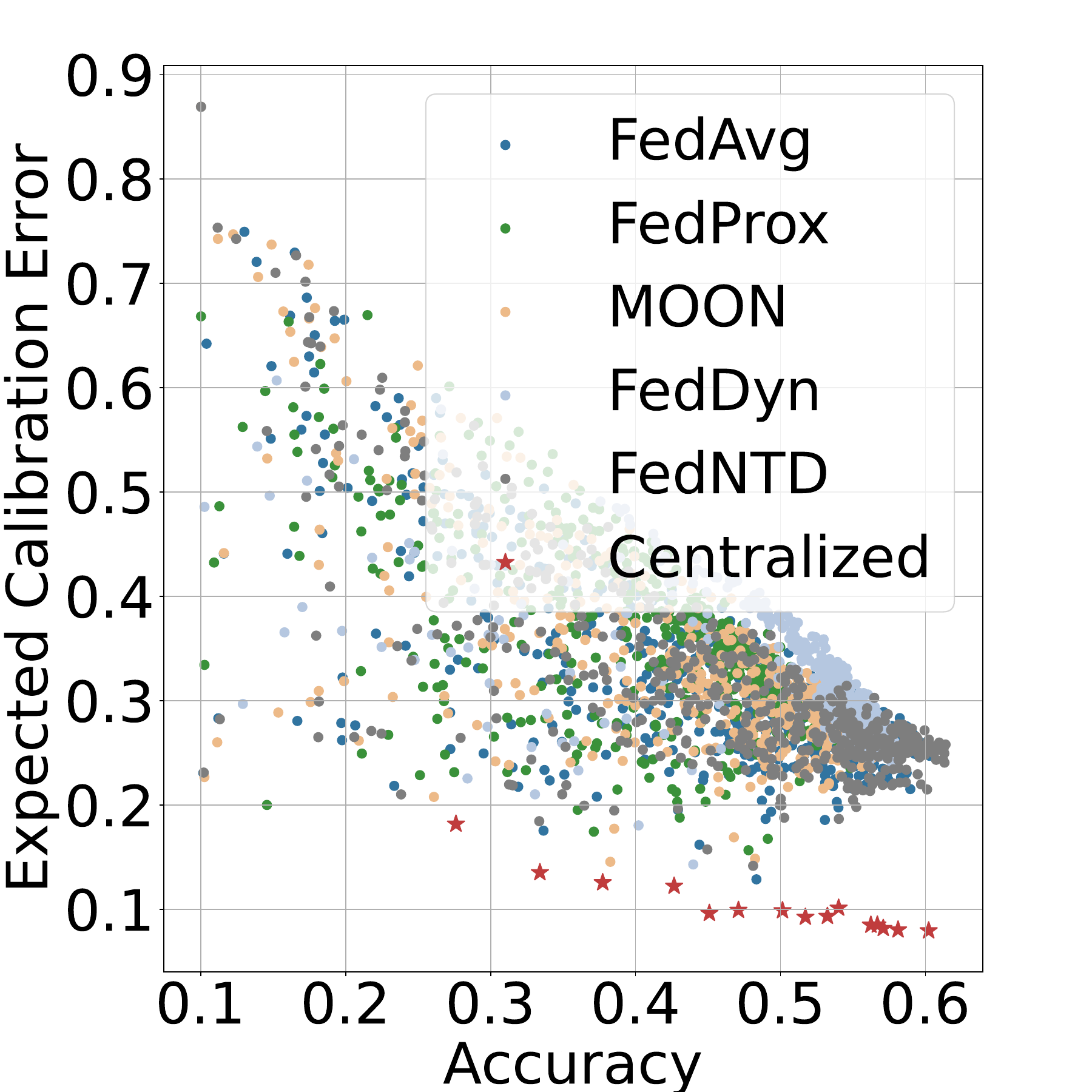}
          \caption{F-ECE vs. ACC}
          \label{fig3a}
    \end{subfigure}
    \begin{subfigure}{0.2\textwidth}
        \includegraphics[width=\textwidth]{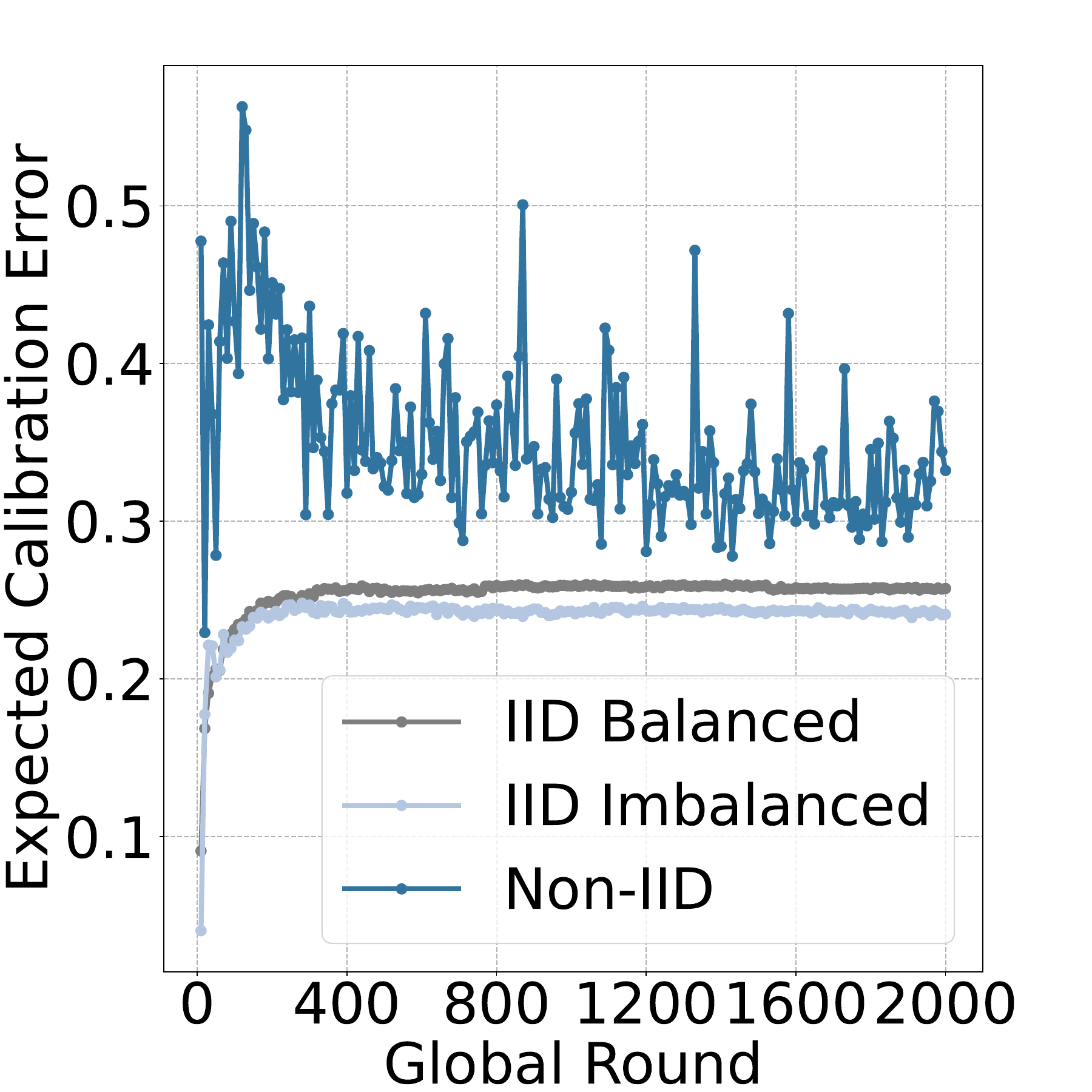}
          \caption{Data Quantity}
          \label{fig3b}
    \end{subfigure}
    \begin{subfigure}{0.2\textwidth}
        \includegraphics[width=\textwidth]{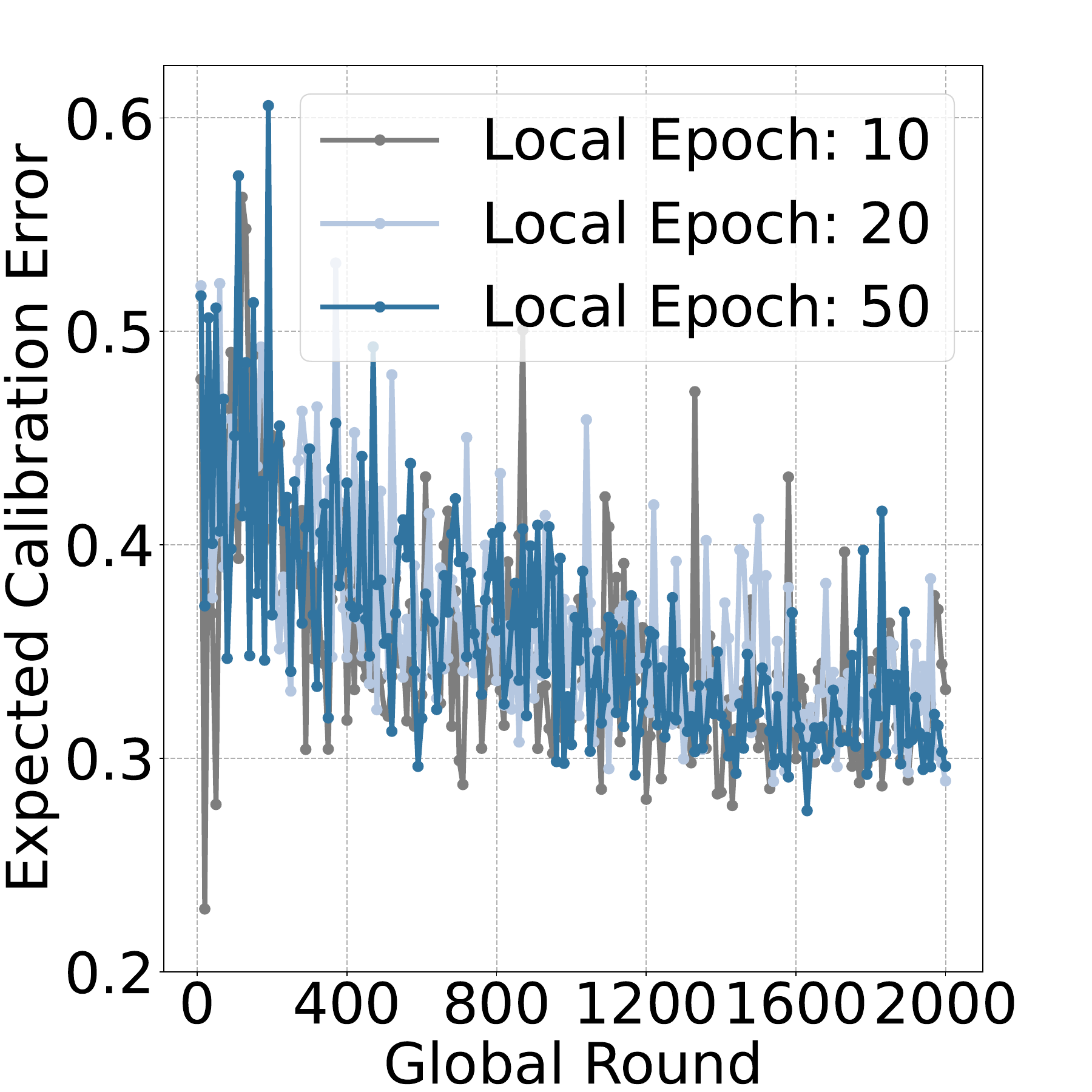}
          \caption{Local Epoch}
          \label{fig3c}
    \end{subfigure}
    \begin{subfigure}{0.2\textwidth}
        \includegraphics[width=\textwidth]{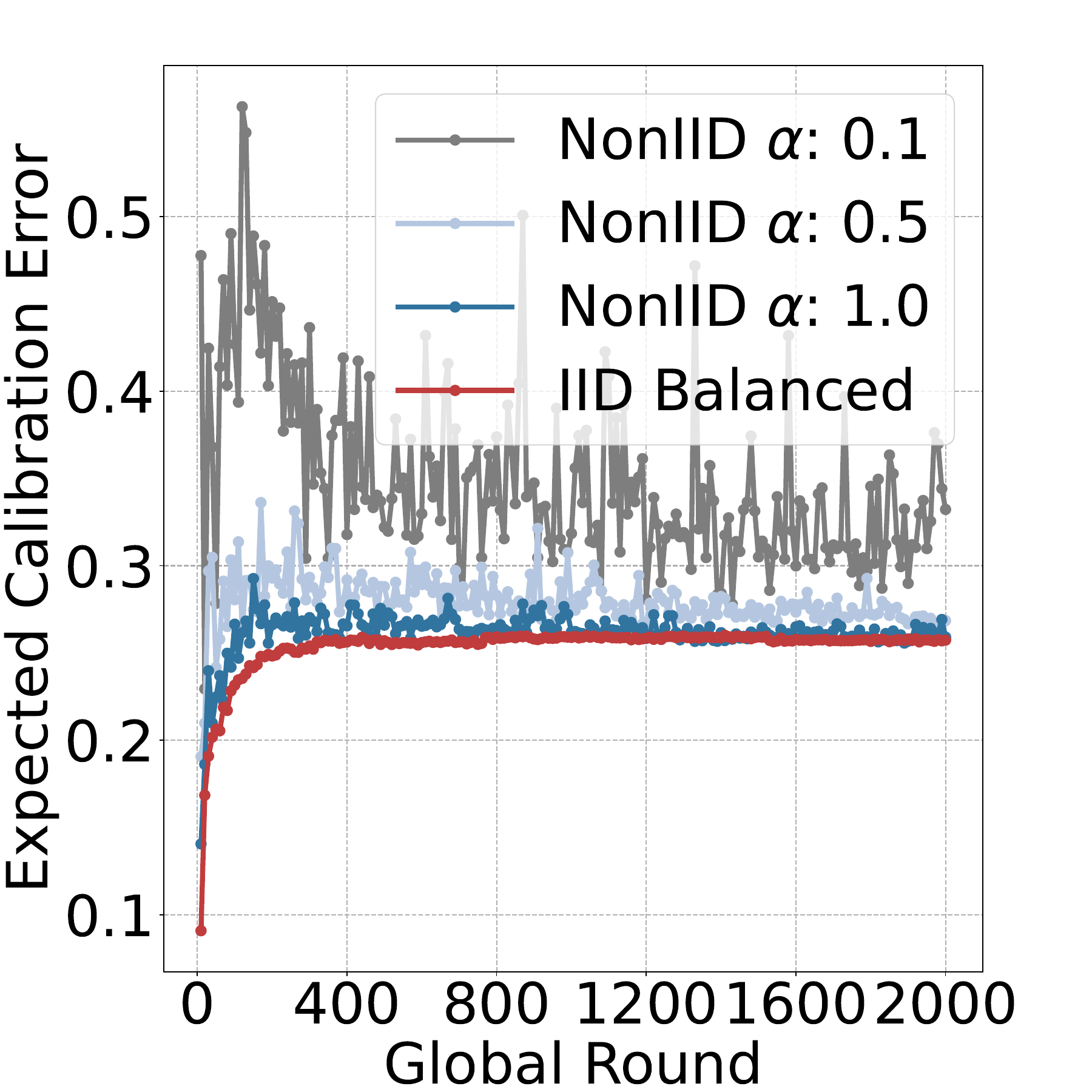}
          \caption{Non-IID Severity}
          \label{fig3d}
    \end{subfigure}
    \begin{subfigure}{0.2\textwidth}
        \includegraphics[width=\textwidth]{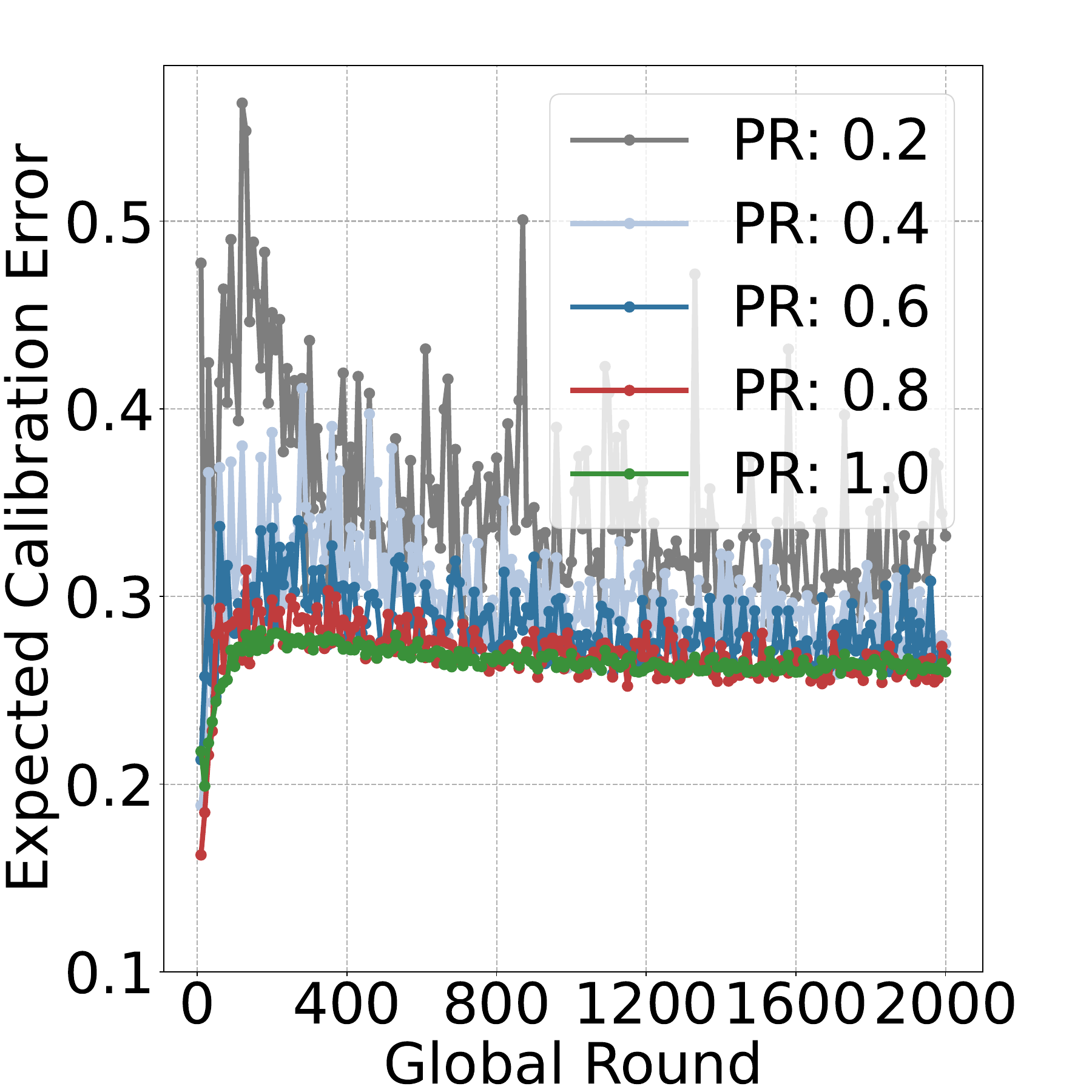}
          \caption{Participation Ratio\\Non-IID Data}
          \label{fig3e}
    \end{subfigure}
    \begin{subfigure}{0.2\textwidth}
        \includegraphics[width=\textwidth]{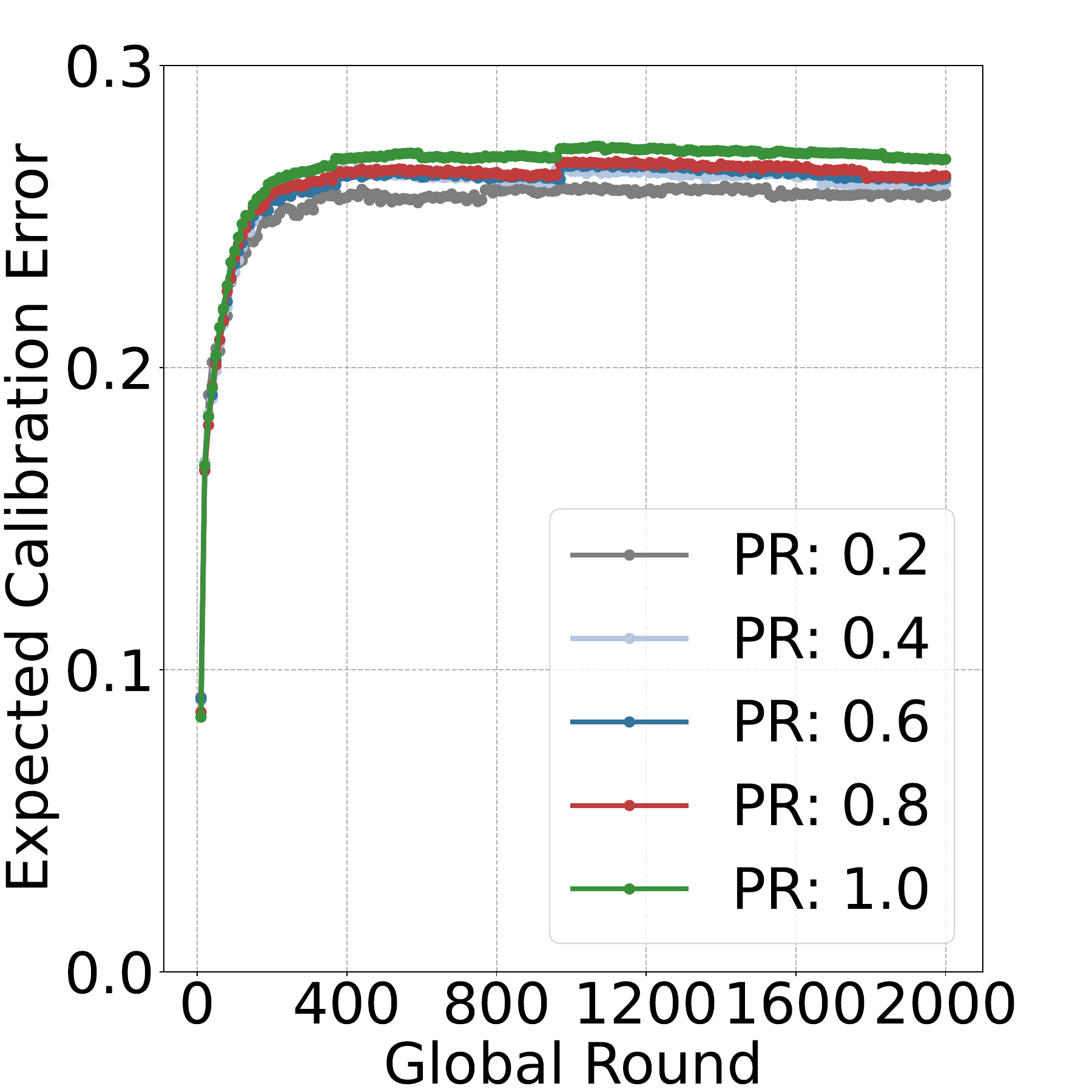}
          \caption{Participation Ratio\\IID Data}
          \label{fig3f}
    \end{subfigure}
    \caption{{Impact Factors on the Reliability of Generic Federated Model.} We investigate the related impact of data quantity imbalance, local epoch number, Non-IID severity, participation ratio on the reliability of the federated model. As can be seen in (b) and (c), data quantity imbalance and local epoch have trivial impacts on the model reliability. (d) and (e) demonstrate that the Non-IID severity significantly harms the federated model's reliability and the low participation ratio magnifies such impact. (f) further illustrate that the participation ratio doesn't affect the reliability in IID data.}
    \label{fig3}
\end{figure}

\subsection{Evaluation Metrics of Model Uncertainty}
\textbf{Expected Calibration Error(ECE)} measures the discrepancy between prediction probability and empirical accuracy, providing an important tool to assess model calibration \cite{naeini2015obtaining}. \\
\textbf{Negative Log-Likelihood(NLL)} is a proper scoring rule for measuring the accuracy of predicted probabilities and evaluating the quality of uncertainty \cite{ovadia2019can}.\\
\textbf{Entropy} of the predictive distribution is a common metric to evaluate the model's reliability when facing OOD data \cite{ovadia2019can}. The histogram of the predictive entropy is always used to compare the uncertainty quality.

\subsection{Related Work}
\textbf{Uncertainty Estimation}. Uncertainty estimation is the most common way to improve a model's reliability in practical scenarios. It aims to produce the accurate uncertainty or confidence of the given sample, reflecting the possibility of fault judgment and indicating whether the sample is out of the knowledge scope of the model. The most common approach for uncertainty estimation is using softmax output in the last layer. However, it is always overconfident in modern neural networks\cite{guo2017calibration}. Existed uncertainty estimation methods can be roughly divided into Bayesian and Non-Bayesian methods. Bayesian methods \cite{louizos2016structured, riquelme2018deep} always involve the computation of the posterior distribution of parameters, which is computationally intractable due to numerous non-linear operations in the network forward passes. A variety of approximation methods has been developed, including variational inference \cite{graves2011practical}, Monte Carlo Markov Chain \cite{welling2011bayesian}, MCDropout \cite{gal2016dropout}, etc. Though the Bayesian model could estimate model uncertainty through parameter posterior distribution, most of these methods can not be applied to modern networks due to their complexity. Different from Bayesian methods, ensemble-based methods do not put a distribution over model parameters \cite{lakshminarayanan2017simple}. Instead, they train several independent models with different random initializations on the same dataset. $\Delta$-UQ \cite{deltauq} utilizes the NTK \cite{NTK} to approximate the training procedure of the ensembles to estimate uncertainty.  During inference, the ensemble will average the outputs to form the prediction, which introduces additional computation costs. Different from the ensemble-based methods, EDL \cite{EDL} and DUQ \cite{DUQ} can estimate uncertainty in a single forward pass.

\textbf{Federated Learning.} Current SOTA federated learning methods can be divided into generic federated learning (G-FL) methods and personalized federated learning (P-FL) methods. The typical algorithm of the former G-FL methods is the FedAvg \cite{mcmahan2017communication}, which aims to train a single generic model for all clients. However, heterogeneity greatly hinders the performance and the generalization ability of the federated global model. To tackle this issue, numerous methods have been proposed from the perspectives of local drift mitigation \cite{feddc, fedprox}, gradient revision \cite{FedAvgM, FedDyn}, knowledge preservation \cite{ensembleDis, FedNTD}, etc. Though improvement has been achieved, there is still a significant performance gap between the federated model and the centralized training model. Different from the single generic model, P-FL methods expect to train each client in a personalized model to fit their unique data distribution. It preserves the distribute-and-aggregate schema of the classic federated learning and regularizes the personalized model with generic information. The idea of the P-FL is first proposed in \citet{FedMultitask}, and further formally extended by \citet{FedPLayer}. Inspired by these pioneer work, a great number of P-FL methods with different strategies have been proposed, such as fine-tuning the global model \cite{Per-FedAvg}, splitting and fine-tuning the client-specific head \cite{FedRep}, learning additional personalized models \cite{pFedMe, ditto}, aggregating with personalized strategies \cite{FedAMP, FedFOMO, FedALA}.

\begin{figure}[t!]
\captionsetup[subfigure]{justification=centering}
    \centering
      \begin{subfigure}{0.2\textwidth}
        \includegraphics[width=\textwidth]{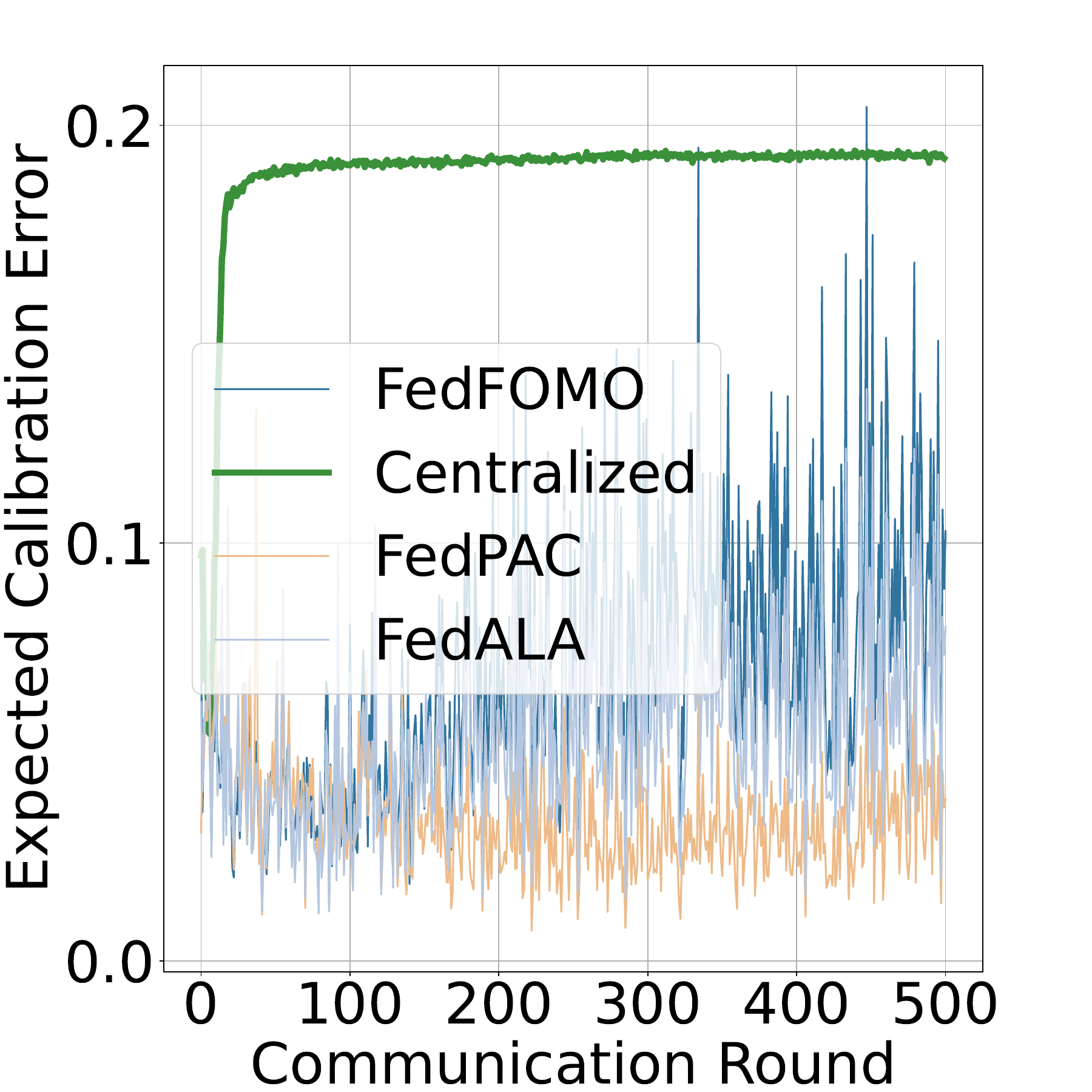}
          \caption{F-ECE}
          \label{fig4a}
      \end{subfigure}
      \begin{subfigure}{0.2\textwidth}
        \includegraphics[width=\textwidth]{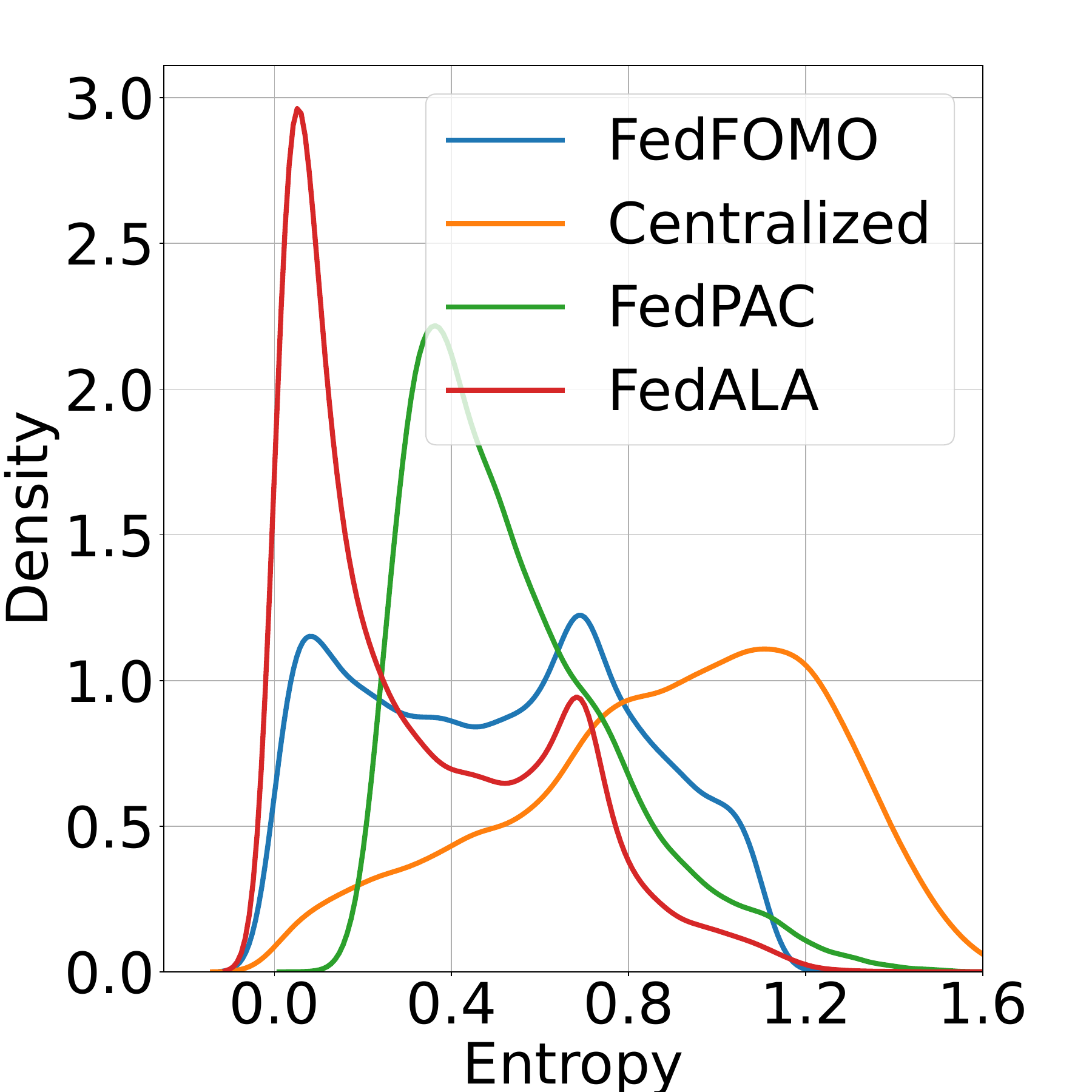}
          \caption{ Predictive Entropy}
          \label{fig4b}
      \end{subfigure}
    \caption{{Reliability of Personalized Federated Models.} (a) F-ECE of personalized federated models compared with centralized training model on the in-domain test dataset. (b) Histogram of predictive distribution entropy on OOD test data. Compared with the centralized training model, personalized models are more calibrated, while still exhibiting lower uncertainty when faced with OOD samples.}
    \label{fig4}
\end{figure}

\section{Reliability of the Federated Models}
To explore how the federated optimization influences the reliability of the obtained model, we conduct a systematic experiment on different SOTA federated methods. We use the Dirichlet distribution $p \sim \operatorname{Dir}(\alpha)$ to assign the proportion of class $k$ from Cifar10 to each client. The total client number is 20, and the default local epoch is set to 10. To get a more comprehensive view of the reliability, we use different SOTA federated frameworks to train federated models, including FedAvg \cite{mcmahan2017communication}, FedProx \cite{fedprox}, FedDyn \cite{FedDyn}, MOON \cite{li2021model}, FedNTD \cite{FedNTD} for G-FL, and FedFOMO \cite{FedFOMO}, FedALA \cite{FedALA}, FedPAC \cite{FedPAC} for P-FL. 


\subsection{Generic Federated Models are Not Reliable}
To investigate the influence of federated optimization on the model's reliability, we first consider exploring the calibration on in-domain test data by measuring the F-ECE of the obtained federated model. The larger the F-ECE, the more miscalibrated and thus less reliable the model is. To unify the comparison between G-FL and P-FL methods, we propose to measure the federated expected calibration error for both generic and personalized federated models.

\begin{definition}
Consider a federated learning framework with $N$ clients for a $K$-class classification problem. Each client $c_i$ has its own test $\mathcal{D}_i^t = \left\{\left(\mathbf{x}_j^i, \mathbf{y}_j^i\right)\right\}_{j=1}^{\hat n_i}$ and model parameter $\boldsymbol{\theta}_i$. Given the partitions $0=l_0^i<\ldots<l_{S}^i=1$, the federated expected calibration error (F-ECE) is defined as:

\begin{equation}
\operatorname{F-ECE} = \sum_{i=1}^{N}{\sum_{s=1}^{S}{\frac{\left|\hat{B}_s^i\right|}{\sum_{i=1}^{N}{\hat{n}_i}}}\left|\operatorname{conf}_s^i - \operatorname{acc}_s^i \right|}
\end{equation}
\begin{equation}
     \operatorname{conf}_s^i = \frac{{\sum_{j \in \hat{B}_s^i}{\hat p(\mathbf{x}_j^i|\boldsymbol{\theta}_i)}}}{{\left|\hat{B}_s^i \right|}}
\end{equation}
\begin{equation}
    \operatorname{acc}_s^i = \frac{{\sum_{j \in \hat{B}_s^i}{\mathbf{1}\left(\mathbf{\hat y}\left(\mathbf{x}_j^i|\boldsymbol{\theta}_i\right) = \mathbf{{y}}_{j}^{i}\right)}}}{{\left|\hat{B}_s^i \right|}}
\end{equation},
where $\hat{B}_s^i = \left\{j \mid l_{s-1}^i < \hat{p}(\mathbf{x}_j^i|\boldsymbol{\theta}_i) \leq l_s^i; \forall \left(\mathbf{x}_j^i, \mathbf{y}_j^i\right) \in \mathcal{D}_i^t \right\} $, $\mathbf{\hat y}\left(\mathbf{x}_j^i|\boldsymbol{\theta}_i\right)$ and $\hat p(\mathbf{x}_j^i|\boldsymbol{\theta}_i)$ are the assigned class and confidence of $\mathbf{x}_j^i$ predicted by model $f\left(\cdot, \boldsymbol{\theta}_i\right)$. 
\end{definition}

\begin{figure}[t!]
\captionsetup[subfigure]{justification=centering}
    \centering
      \begin{subfigure}{0.2\textwidth}
        \includegraphics[width=\textwidth]{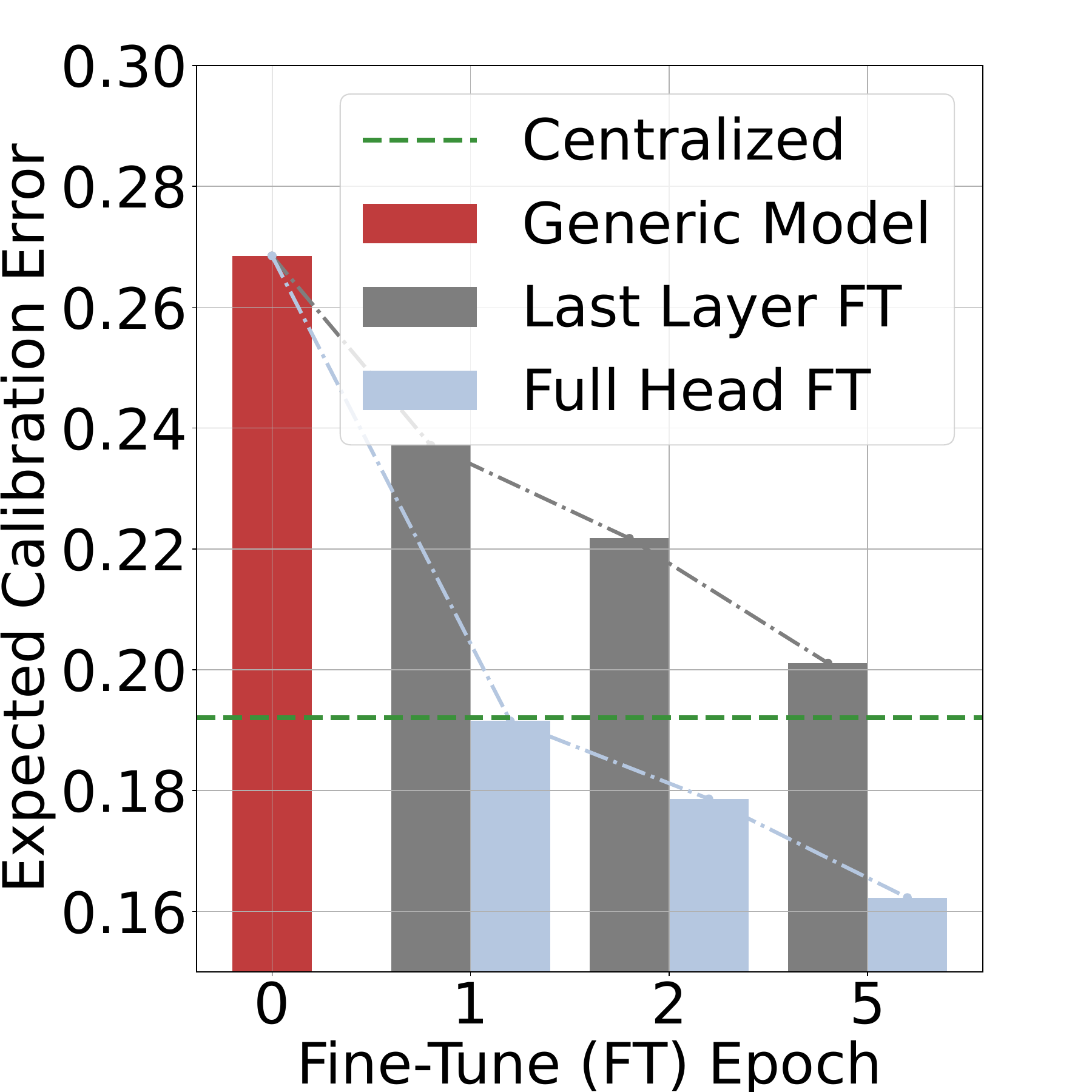}
          \caption{F-ECE}
          \label{FT1}
      \end{subfigure}
      \begin{subfigure}{0.2\textwidth}
        \includegraphics[width=\textwidth]{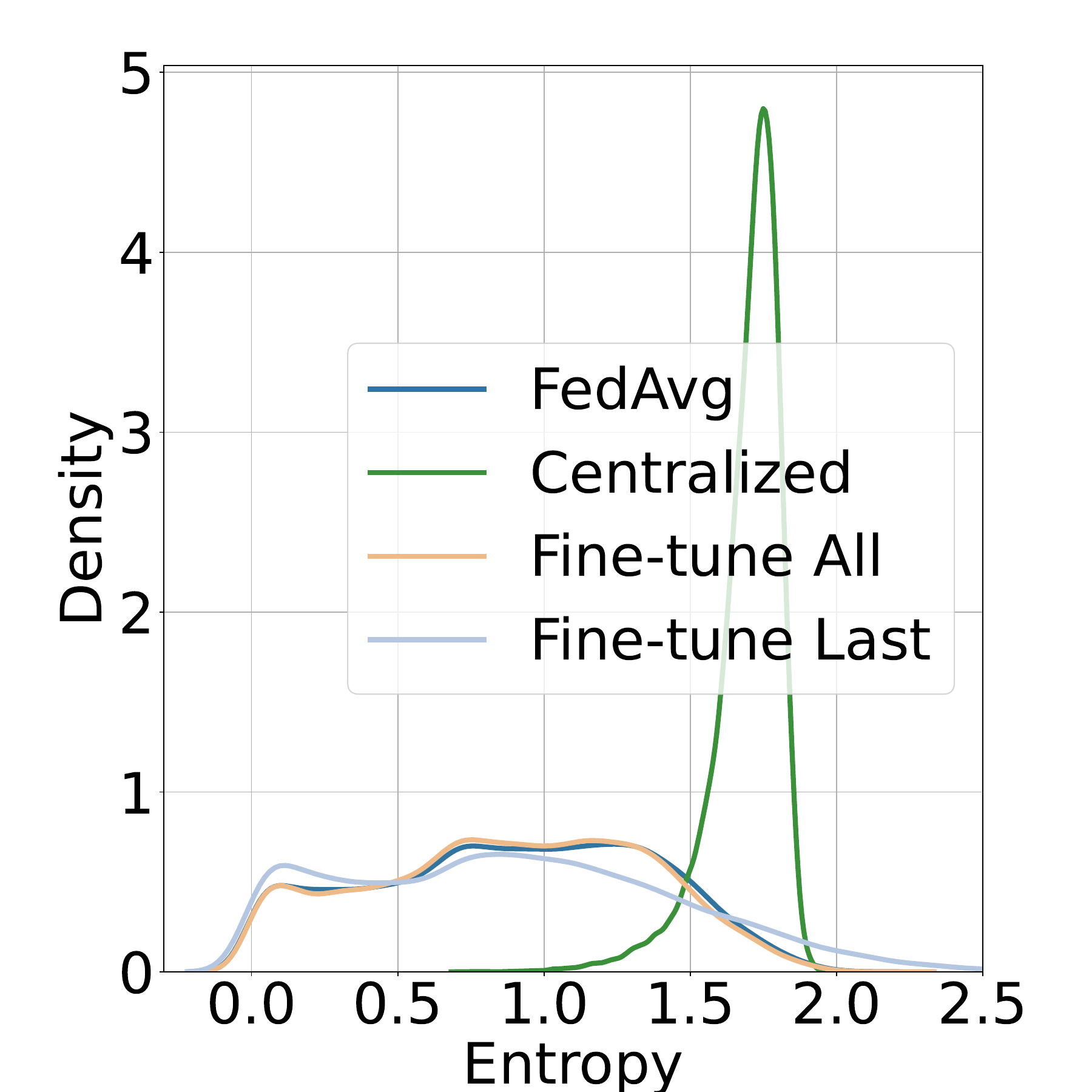}
          \caption{Predictive Entropy}
          \label{FT2}
      \end{subfigure}
    \caption{{Influence of Head Fine-tuning.} (a) Bar diagram of F-ECE before/after head fine-tuning.  (b) Histogram of predictive entropy on OOD samples. The model achieves lower ECE than the centralized model (green dash line) after only 1 round head fine-tuning, while the uncertainty to OOD samples remains unreliable.} 
    \label{GFL-Finetune-Fig}
\end{figure}

\begin{figure*}[t!]
    \centering
    \includegraphics[width = 0.79\textwidth]{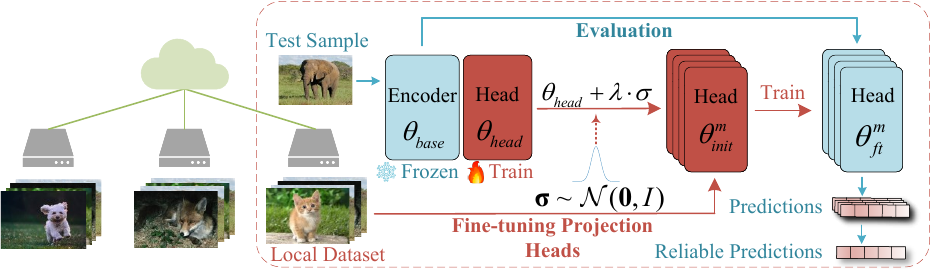}
    \caption{The Overall Framework of Proposed APH Method. Treating the origin parameter of the projection head as prior, APH samples permutation from Gaussian distribution and obtains multiple projection heads with different initializations, which are further fine-tuned with various learning rates on the local dataset. The reliable prediction is then obtained by head averaging.}
    \label{mainfig}
\end{figure*}

We train the generic federated models with different SOTA methods on heterogeneous Cifar10 and explore their F-ECE on in-domain test data and predictive distribution entropy on OOD test data. The experimental results are displayed in Fig. \ref{fig2}. Experimental results demonstrate that all models obtained from G-FL methods demonstrate a higher F-ECE on in-domain test data and lower predictive entropy on OOD test data than the traditional centralized training model, which indicates significant overconfidence and severe unreliability problems of generic federated models. Figure \ref{fig3a} further demonstrates that such loss of reliability is not caused by the degradation of the accuracy.

We further explore the related impact factors on the reliability of generic federated models. Without loss of generality, we choose FedAvg as the federated framework in the following experiments. We investigate the F-ECE under different federated settings, e.g. local epoch number, partial participation ratio, data quantity imbalance, and severity of Non-IID distribution. The experimental results are shown in Fig. \ref{fig3}. As demonstrated in Fig. \ref{fig3b}, \ref{fig3c} and \ref{fig3e}, data quantity, local epochs, as well as partial participation under IID data have a trivial impact on the reliability of the generic federated model. However, the Non-IID severity significantly harms the federated model's reliability, and the low participation ratio magnifies such impact (See Fig. \ref{fig3d} and \ref{fig3e}).

\subsection{Personalized Federated Models are Not Solutions}
We further turn our gaze to the personalized federated models. Similarly, we train the personalized federated model utilizing SOTA personalized federated methods on Cifar10 with practical settings. The F-ECE on in-domain test data and the predictive entropy are evaluated to measure the reliability of the obtained personalized federated models. We report the experimental results of FedPAC \cite{FedPAC}, FedALA \cite{FedALA}, and FedFOMO \cite{FedFOMO} compared with the centralized training schema in Fig. \ref{fig4}. Results demonstrate that the personalized federated model is more calibrated than the centralized training model, while still exhibiting lower uncertainty to OOD test samples, indicating that personalized federated models are not the solution for reliable federated learning.

\subsection{Projection Head Bias is the Primary Cause}
Although the conclusion is depressing, we further conduct an interesting experiment motivated by CCVR \cite{CCVR} and FedRoD \cite{FedRoD}, in which they point out that the classifier of the federated model is biased and local updated models of FedAvg are naturally well-personalized federated models respectively. Our experiment revolves around the single question: \textbf{Whether the projection heads of federated models are the main cause of the degradation of reliability?}  To answer this question, we fine-tune the projection head of local models by utilizing their local dataset while keeping the feature extractors frozen. After the fine-tuning of the projection head, we further evaluate its F-ECE on in-domain test data and predictive entropy on OOD data. Results are displayed in Fig. \ref{GFL-Finetune-Fig}.

Specifically, we adopt two strategies that fine-tune the last fully connected layer and the whole projection head respectively. As demonstrated in Fig. \ref{GFL-Finetune-Fig}, the generic federated model achieves lower F-ECE than the centralized training model with the same accuracy after only one round of full-head fine-tuning. It further gets a significant F-ECE decrease after only 5 fine-tuning rounds. Oppositely, the result of the last layer fine-tuning is not satisfying but still gets a decrease on F-ECE. Moreover, we observe that the histograms of predictive entropy on the OOD samples of the head fine-tuned models are almost the same as the generic model's, indicating nearly no improvement in the uncertainty estimation obtained from the fine-tuning of the projection head. 

Such an observation validates our conjecture: the biased projection head is one of the main causes of the degradation of model reliability. Intuitively, the data heterogeneity and partial participation strategies always lead to inconsistency of the averaged gradients, causing severe overfitting of projection heads. The success of model calibration and the failure of model uncertainty of the head fine-tuning strategy demonstrate that it is not enough to deal with projection heads alone, parameter diversity must be obtained to actually improve the reliability of the federated model.

\begin{table*}[t!]
  \centering
  \resizebox{0.85\linewidth}{!}{
    \begin{tabular}{cccccccccc}
    \toprule
          & \multicolumn{3}{c}{Cifar10} & \multicolumn{3}{c}{Cifar100} & \multicolumn{3}{c}{Tiny-ImageNet} \\
\cmidrule{2-10}          & Accuracy & F-ECE & NLL   & Accuracy & F-ECE & NLL   & Accuracy & F-ECE & NLL \\
    \midrule
    FedAvg & 0.615 & 0.168 & 1.454 & 0.284 & 0.458 & 5.586 & 0.098 & 0.395 & 6.200 \\
    FedDropout & 0.619 & 0.138 & 1.333 & 0.282 & 0.458 & 6.091 & 0.115 & 0.290  & 5.115 \\
    Client Ensembles & 0.601 & 0.130  & 1.252 & 0.286 & 0.395 & 5.194 & 0.127 & 0.081 & 4.717 \\
    FedAvg + FineTune & 0.784 & 0.073 & 0.823 & 0.323 & 0.426 & 5.094 & 0.239 & 0.275 & 4.459 \\
    FedAvg + APH & \textbf{0.902} & \textbf{0.036} & \textbf{0.311} & \textbf{0.484} & \textbf{0.091} & \textbf{2.747} &    \textbf{0.294}   &  \textbf{0.089}     & \textbf{3.671} \\
    \bottomrule
    \end{tabular}%
    }
    \captionsetup{justification=raggedright,singlelinecheck=false}
    \caption{Effectiveness of APH on model calibration. FedAvg + FineTune considers single-head fine-tuning as a baseline.}
    \label{tab1}%
\end{table*}%

\begin{table*}[t!]
  \centering

  \resizebox{0.85\linewidth}{!}{
    \begin{tabular}{cccccccccccc}
    \toprule
    \multicolumn{2}{c}{Method} & \multicolumn{2}{c}{FedProx} & \multicolumn{2}{c}{FedDyn} & \multicolumn{2}{c}{FedNTD} & \multicolumn{2}{c}{FedALA} & \multicolumn{2}{c}{FedFOMO} \\
    \multicolumn{2}{c}{With/Without APH} & Without & With  & Without & With  & Without & With  & Without & With  & Without & With \\
    \midrule
    \multirow{3}[2]{*}{Cifar10} & Acc   & 0.556 & \textbf{0.897} & 0.559 & \textbf{0.861} & 0.553 & \textbf{0.900} & 0.894 & \textbf{0.899} & 0.877 & \textbf{0.881} \\
          & F-ECE & 0.226 & \textbf{0.066} & 0.102 & \textbf{0.034} & 0.263 & \textbf{0.033} & 0.065 & \textbf{0.047} & 0.090  & \textbf{0.064} \\
          & NLL   & 1.662 & \textbf{0.344} & 9.333 & \textbf{0.464} & 2.028 & \textbf{0.326} & 0.460  & \textbf{0.325} & 0.916 & \textbf{0.397} \\
    \midrule
    \multirow{3}[2]{*}{Cifar100} & Acc   & 0.197 & \textbf{0.275} & 0.225 & \textbf{0.486} & 0.298 & \textbf{0.485} & 0.416 & \textbf{0.427} & \textbf{0.315} & 0.303 \\
          & F-ECE & 0.41  & \textbf{0.078} & 0.616 & \textbf{0.102} & 0.449 & \textbf{0.128} & 0.396 & \textbf{0.090} & 0.337 & \textbf{0.114} \\
          & NLL   & 5.215 & \textbf{3.679} & 10.602 & \textbf{3.800} & 5.444 & \textbf{2.823} & 4.833 & \textbf{3.135} & 3.452 & \textbf{2.918} \\
    \midrule
    \multirow{3}[2]{*}{Tiny-ImageNet} & Acc   & 0.083 & \textbf{0.166} & 0.089 & \textbf{0.121} & 0.114 & \textbf{0.338} & 0.293 & \textbf{0.298} & \textbf{0.215} & 0.210 \\
          & F-ECE & 0.238 & \textbf{0.108} & 0.602 & \textbf{0.073} & 0.351 & \textbf{0.149} & 0.326 & \textbf{0.105} & 0.201 & \textbf{0.112} \\
          & NLL   & 5.439 & \textbf{4.517} & \textbf{10.496} & 13.769 & 5.678 & \textbf{3.149} & 4.098 & \textbf{3.797} & 4.034 & \textbf{3.742} \\
    \bottomrule
    \end{tabular}
    }

\captionsetup{justification=raggedright,singlelinecheck=false}
  \caption{Compatibility of APH with generic and personalized federated methods across prominent federated benchmarks. }
  \label{tab2}%
\end{table*}%

\section{Assembling Projection Heads to Make Federated Model Reliable Again}
As discussed before, fine-tuning the projection head is not sufficient to encounter the overconfidence problem of deep neural networks, leading to a higher accuracy but still bad reliability performance. While single fine-tuning is struggling, we borrow the idea from the deep ensemble \cite{lakshminarayanan2017simple} to introduce the parameter diversity into the model inference by fine-tuning multiple projection heads with different initialization.

 We now propose Assembled Projection Heads (APH). Specifically, for model $f(\cdot;\boldsymbol{\theta}_{i}^{\mathcal{R}})$ of the client $c_i$ which is obtained by either generic or personalized federated methods after $\mathcal{R}$ rounds, APH splits the model into two parts, i.e. feature extractor $\phi(\cdot; \boldsymbol{\theta}_{i,\text{base}}^{\mathcal{R}})$and projection head $h(\cdot; \boldsymbol{\theta}_{i,\text{head}}^{\mathcal{R}})$. Freezing the feature extractor, APH treats the parameter of the projection head as the prior, and further samples from the Gaussian distribution to permute the prior to get multiple initialized parameters. The initialized parameter of the projection head of client $c_i$ is given as:
$$
\boldsymbol{\theta}_{i, \text{init}}^{\mathcal{R}} = \boldsymbol{\theta}_{i,\text{head}}^{\mathcal{R}} + 10 ^ {\lambda} \cdot \mathbf{\sigma}
$$, where $\sigma \sim \mathcal{N}\left(\mathbf{0}, \mathbf{I}\right)$, $\lambda$ is the hyper-parameter which controls the magnitude of $\mathbf{\sigma}$, $\mathbf{I}$ is an identity matrix whose trace equals the parameter number in $\boldsymbol{\theta}_{i, \text{head}}$. After getting the $M$ initialized heads, APH fine-tunes these projection heads on the local dataset for $E_{h}$ epochs respectively with various learning rates to get the fine-tuned projection heads $\{\boldsymbol{\theta}_{i,\text{head}}^{\mathcal{R}, m}\}_{m=1}^{M}$. The learning rate $\beta$ is sampled from an uniform distribution $\beta \sim [\beta_{l}, \beta_{u}]$. The final prediction of $\mathbf{x}$ in client $c_i$ is given as
$
\hat{\mathbf{y}} = \frac{1}{M}\sum_{m=1}^{M}{h\left(\phi(\mathbf{x}; \boldsymbol{\theta}_{i;\text{base}}^{\mathcal{R}}); \boldsymbol{\theta}_{i,\text{head}}^{\mathcal{R}, m}\right)}.
$

The proposed APH method can be easily applied to most generic and personalized federated frameworks to improve the model's reliability. Different from ensemble-based methods which require $N_{\text{inf}} \times$ inference time for $N_{\text{inf}}$ runs, APH only involves additional computation in multiple projection heads, which is a quite small fraction in the total computation cost of inference. For large models such as ResNet-50, the computational overhead of the projection head is only around 0.3\% of the whole inference process. Thus even for the APH method which possesses 100 projection heads, the additional computation cost is still less than 30\%.

\begin{figure*}[t!]
\captionsetup[subfigure]{justification=centering}
    \centering
      \begin{subfigure}{0.25\linewidth}
        \includegraphics[width=\textwidth]{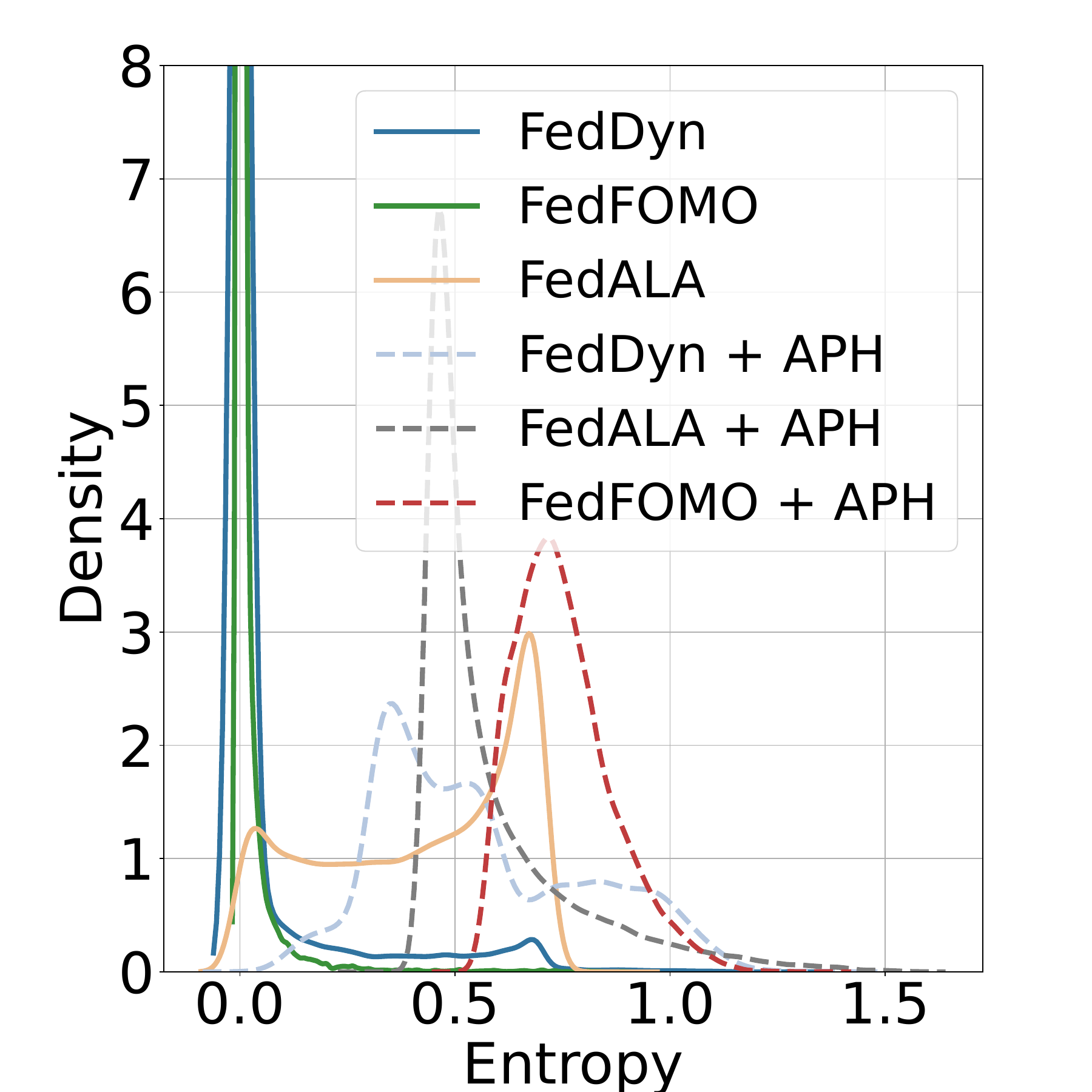}
          \caption{SVHN vs. Cifar10}
          \label{OOD1}
      \end{subfigure}
      \begin{subfigure}{0.25\linewidth}
        \includegraphics[width=\textwidth]{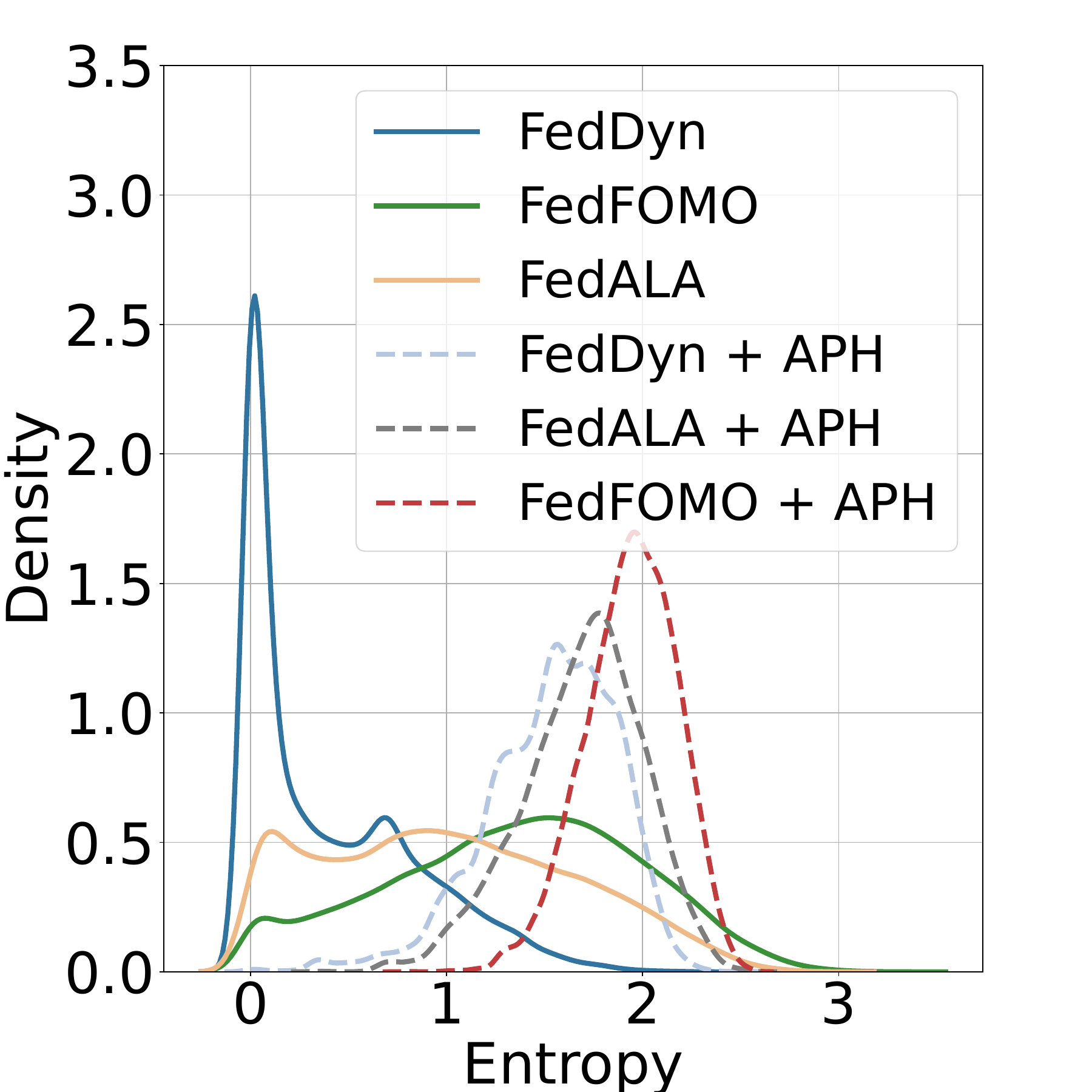}
          \caption{SVHN vs. Cifar100}
          \label{OOD2}
      \end{subfigure}
      \begin{subfigure}{0.25\linewidth}
        \includegraphics[width=\textwidth]{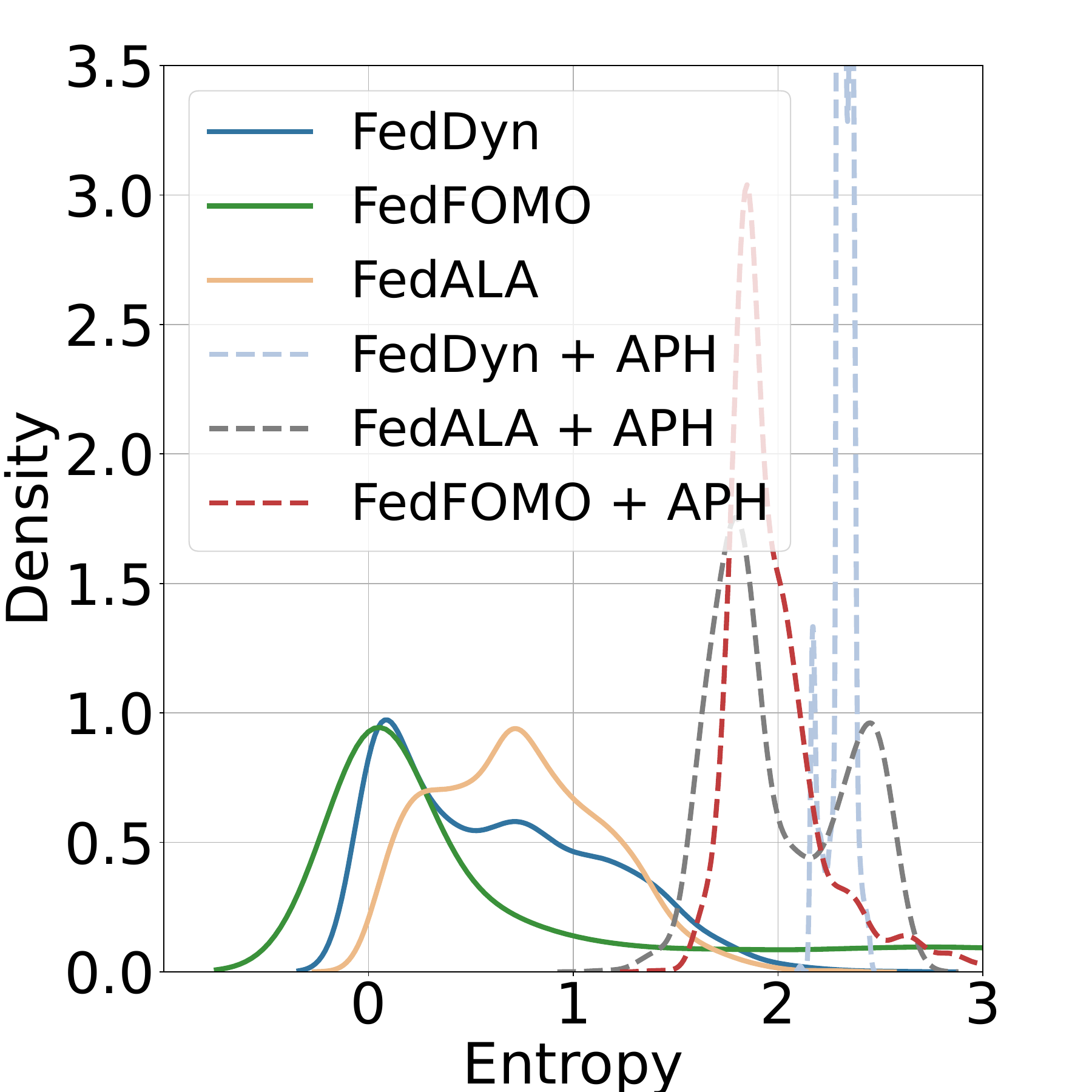}
          \caption{ImageNet-O vs. Tiny-ImageNet}
          \label{OOD3}
      \end{subfigure}
    \caption{Effectiveness of APH in Improving the Model Reliability on Out-Of-Distribution Data.}
    \label{OOD}
\end{figure*}

\section{Experiments}
\subsection{The Setup}
In this section, we will briefly introduce the details and related settings of our experiments. Due to the limited spaces, more experimental settings and results (Ablation study, Robustness on hyper-parameters) can be found in Appendix.\\
\textbf{Dataset.} We conduct experiments on popular federated dataset Cifar10, Cifar100 \cite{krizhevsky2009learning}. To further validate the effectiveness of the proposed APH on large datasets, we also conduct experiments on the Tiny-ImageNet dataset \cite{le2015tiny}. We partition each of the datasets into 10 clients with the default heterogeneous setting. The participation strategy is the same as the experiment in Section 3. For OOD dataset, we use SVHN \cite{netzer2011reading} for Cifar10/100, and ImageNet-O \cite{ImageNet-O} for Tiny-ImageNet.\\
\textbf{Methods.} We evaluate the effectiveness of our method compared with the simple baseline MC-Dropout, deep ensembles of personalized client models. Other calibration and uncertainty estimation methods are not applicable due to the federated training schema (See Appendix for details). To validate the effectiveness and the compatibility of proposed APH with other SOTA federated methods, we also apply APH to SOTA generic federated methods FedProx, FedDyn, FedNTD and personalized methods FedFOMO, FedALA.\\
\textbf{Models.} We train a CNN on the Cifar10 dataset, and further validate the effectiveness on large models by training ResNet-50 on the Cifar100 and Tiny-ImageNet datasets. \\
\textbf{Hyperparameters.} For federated methods, we set the global communication round to 100 for Cifar10/100, and 20 for Tiny-ImageNet. The local epoch number $E$ is set to 10. For the dropout ratio in FedDrop, we select the best result from $\{0.1, 0.2, 0.5\}$. For the $\lambda$ used in APH, we select the $\lambda$ from $\{\mu-0.5, \mu -0.2, \mu, \mu+0.2, \mu+0.5\}$, where $\mu$ is the magnitude of the order of the mean value of the parameter. For the lower bound of learning rate, we set the $\beta_l$ to 0.001. For the upper bound $\beta_u$, we choose the best result from $\{10, 1, 0.1\}$.

\begin{table}[t!]
\centering
  \resizebox{0.80\linewidth}{!}{
    \begin{tabular}{cccccc}
    \toprule
    \multicolumn{2}{c}{\multirow{2}[2]{*}{Parameter}} & \multicolumn{2}{c}{Accuracy} & \multicolumn{2}{c}{F-ECE} \\
    \multicolumn{2}{c}{} & Without & With  & Without & With \\
    \midrule
    \multirow{3}[2]{*}{$\alpha$} & 0.05  & 0.578  & \textbf{0.947 } & 0.123  & \textbf{0.015 } \\
          & 0.1   & 0.598  & \textbf{0.900 } & 0.162  & \textbf{0.041 } \\
          & 0.5   & 0.667  & \textbf{0.763 } & 0.241  & \textbf{0.106 } \\
    \midrule
    \multirow{3}[2]{*}{$\gamma$} & 0.2   & 0.420  & \textbf{0.885 } & 0.280  & \textbf{0.041 } \\
          & 0.6   & 0.597  & \textbf{0.901 } & 0.173  & \textbf{0.038 } \\
          & 1     & 0.598  & \textbf{0.900 } & 0.162  & \textbf{0.041 } \\
    \midrule
    \multirow{3}[2]{*}{$E$} & 10    & 0.598  & \textbf{0.900 } & 0.162  & \textbf{0.041 } \\
          & 20    & 0.606  & \textbf{0.899 } & 0.175  & \textbf{0.041 } \\
          & 40    & 0.592  & \textbf{0.896 } & 0.205  & \textbf{0.042 } \\
    \midrule
    \multirow{3}[2]{*}{$N$} & 10    & 0.598  & \textbf{0.900 } & 0.162  & \textbf{0.041 } \\
          & 50    & 0.569  & \textbf{0.889 } & 0.183  & \textbf{0.056 } \\
          & 100   & 0.588  & \textbf{0.909 } & 0.123  & \textbf{0.054 } \\
    \bottomrule
    \end{tabular}%
    }

    \caption{Evaluation of APH on different federated settings.}
  \label{tab3}%
\end{table}%

\begin{table}[t!]
  \centering
  \resizebox{0.9\linewidth}{!}{
    \begin{tabular}{ccccc}
    \toprule
          & Without & 10    & 50    & 100 \\
    \midrule
    Accuracy & 0.284 & 0.484 & 0.501 & 0.503 \\
    F-ECE   & 0.458 & 0.091 & 0.077 & 0.071 \\
    NLL   & 5.586 & 2.747 & 2.188 & 2.056 \\
    Additional Cost & 0.00\%     & 2.44\% & 12.25\% & 24.49\% \\
    \bottomrule
    \end{tabular}%
    }

    \caption{Computation efficiency of ResNet-50 on Cifar100.}
    \label{tab4}%
\end{table}%

\subsection{Reliability On In-domain Test Data}
For the reliability of in-domain test data, we report the F-ECE and NLL as calibration metrics. Results of unbiased metric (i.e. F-KDE-ECE \cite{F-KDE-ECE}) can be found in Appendix.\\
\textbf{Effectiveness on Calibration.} We first evaluate the effectiveness of APH compared with other available SOTA calibration and uncertainty estimation methods. Experimental results are displayed in the Table \ref{tab1}. As can be seen from the table, the proposed APH significantly reduces the F-ECE of the given model and largely improves its accuracy through multiple projection head assembling. \\
\textbf{Compatibility on SOTA Methods.} We further conduct experiments to validate the compatibility and effectiveness of the proposed APH combined with other SOTA generic and personalized federated methods. We display the experimental results in Table \ref{tab2}. Experimental results demonstrate that the proposed APH can be seamlessly integrated into SOTA federated methods and still performs well in accuracy improvement and model calibration.\\
\textbf{Robustness on Federated Hyperparameters.} We now validate the robustness of APH under the various federated settings. We mainly focus on the crucial hyper-parameters, i.e. client participation ratio $\gamma$, the severity level of data heterogeneity $\alpha$, local epoch $E$, and client number $N$. We report the experimental results in Table \ref{tab3}. As can be seen from the table, APH is robust to federated hyper-parameters.

\subsection{Reliability On Out-Of-Distribution Test Data}
In this section, we evaluate the effectiveness of APH in improving model reliability on the OOD dataset. For clarity, we here report the result of the first client. Detailed results of all clients with various federated methods can be found in the Appendix. As can be seen in Fig. \ref{OOD}, APH can significantly improve the uncertainty level of both generic and personalized federated methods on various datasets, showing its effectiveness in improving reliability to OOD samples. We also evaluate the effectiveness of APH under domain shifts on Cifar-C \cite{Cifar-CP}  in Appendix.

\subsection{Analysis on Computation Efficiency}
Computation cost is always the key concern in uncertainty estimation. We here conduct experiments to validate the computation efficiency of APH. We set the various numbers of projection heads and report the results in Table \ref{tab4}. The computation cost is calculated using floating point operation numbers. As demonstrated in the table, APH achieves significant improvement with only 10 heads. Even for 100 heads, APH requires less than $30\%$ additional computational cost.

\section{Conclusion}
In this paper, we conduct a systematic experiment about the reliability of federated models. We uncover the fact that federated models are not reliable under heterogeneous data. We further investigate the impact factors and point out biased projection head is one of the main causes of reliability degradation. Motivated by the observation, we propose APH, a lightweight but effective uncertainty estimation framework for federated models. By treating the existing projection head parameters as priors, APH randomly samples multiple initialized parameters of projection heads from the prior and further performs targeted fine-tuning on locally available data under varying learning rates. Such a head ensemble introduces parameter diversity into the deterministic model, producing reliable predictions via head averaging.  Experiments conducted on Cifar10, Cifar100, and Tiny-ImageNet validate the efficacy of APH in improving reliability by model calibration and uncertainty estimation.  

\section*{Acknowledgements}
This work is supported in part by the National Key Research and Development Program of China under Grant 2020AAA0109602;  by the National Natural Science Foundation of China under Grant 62306074.

\bibliography{aaai24}

\end{document}